\pdfoutput=1
\documentclass[11pt]{article}

\usepackage[preprint]{acl}

\usepackage{times}
\usepackage{latexsym}

\usepackage[section]{placeins}

\usepackage[T1]{fontenc}
\usepackage[utf8]{inputenc}

\usepackage{microtype}

\usepackage{inconsolata}

\usepackage{graphicx}
\usepackage{pdfpages}
\usepackage{subcaption}
\usepackage{enumitem}

\usepackage{hyperref}
\usepackage{adjustbox}
\usepackage{array}
\usepackage{booktabs}
\newcolumntype{P}[1]{>{\centering\arraybackslash}p{#1}}

\title{On Recipe Memorization and Creativity in Large Language Models:\\\emph{Is Your Model a Creative Cook, a Bad Cook, or Merely a Plagiator?}}

 \author{Jan Kvapil, Martin Fajčík\\
   Brno University of Technology, Brno, Czech Republic \\ \texttt{xkvapi19@stud.fit.vutbr.cz},
   \texttt{martin.fajcik@vut.cz}}

\begin{document}
\maketitle
%We aim to study memorization, creativity, and non-sense during recipe generation by creating a novel automatized system.
% We created an annotation scheme to analyse the similarities and potentially memorised ingredients and steps in cooking recipes. 
%Our goals are (i) to analyse potential approaches to automatize human evaluation approach in order to scale our study to hundreds of recipes.
% automatic annotation of recipes with the goal of analysing percentages of potentially memorised recipes in various LLMs
%and (ii) to analyse memorization and creativity in LLMs using a small high-quality set of human judgements. We aim to study hallucinations, memorization and creativity during recipe generation by creating a new, automatized system. 

%We created a system that allows us to quantify the performance of LLMs against human annotations. 

%The results from this pipelined system

\begin{abstract}
This work-in-progress investigates the memorization, creativity, and nonsense found in cooking recipes generated from Large Language Models (LLMs). 
% Targets
Precisely, we aim (i) to analyze memorization, creativity, and non-sense in LLMs using a small, high-quality set of human judgments and (ii) to evaluate potential approaches to automate such a human annotation in order to scale our study to hundreds of recipes.
% Methods and Contributions
To achieve (i), we conduct a detailed human annotation on $20$ preselected recipes generated by LLM (Mixtral), extracting each recipe's ingredients and step-by-step actions to assess which elements are memorized---i.e., directly traceable to online sources possibly seen during training---and which arise from genuine creative synthesis or outright nonsense. We find that Mixtral consistently reuses ingredients that can be found in online documents, potentially seen during model training, suggesting strong reliance on memorized content.

To achieve aim (ii) and scale our analysis beyond small sample sizes and single LLM validation, we design an ``LLM-as-judge'' pipeline that automates recipe generation, nonsense detection, parsing ingredients and recipe steps, and their annotation. 
For instance, comparing its output against human annotations, the best ingredient extractor and annotator is Llama~3.1+Gemma~2~9B, achieving up to $78\,\%$ accuracy on ingredient matching. This automated framework enables large‐scale quantification of memorization, creativity, and nonsense in generated recipes, providing rigorous evidence of the models’ creative capacities.

%\todo{Přidat výsledky práce, co jsme se zatím dozvěděli, připsat I informace o automatické anotaci}
\end{abstract}

\section{Introduction}
Cooking recipes can be thought of as algorithms or programs. There are (i) ingredients that are needed to cook a recipe---prerequisites needed to execute an algorithm, as well as (ii) steps that need to be followed to achieve the requested result. These ingredients and steps may be memorized by the Large Language Models (LLMs) and may not be their own invention. Past work showed that training data can be extracted from LLMs~\cite{nasr2023scalableextractiontrainingdata} and that LLMs can assist with cooking~\cite{chan2023mangomangoletlettuce}, albeit with low human preference~\cite{diallo-etal-2024-pizzacommonsense}.

\textbf{We pose the following research questions}: \emph{Are~these recipes any good? Did the model come up with them? And didn't the model serve us a recipe that is plagiarized without ever crediting the original author?} 

In our study, we ask whether the recipes generated by LLMs are \emph{creative}, \emph{plagiarized (memorized)}\footnote{Under Czech implementation of EU copyright law (§ 39d), a language model may infringe copyright if it reproduces expressive elements of a protected work outside of research use, and the rightholder has explicitly prohibited use for training (e.g., via \texttt{robots.txt}). Since it's often unclear whether a recipe qualifies as a copyright-protected work, we define \emph{potential plagiarism} as a model generating a recipe with ingredients and steps closely matching a publicly available one(s). This does not imply legal infringement but serves as a practical proxy. For brevity, we sometimes call this ``plagiarized''. 
See Appendix \ref{legal_perspective} for deeper discussion.
}, 
or simply \emph{bad (nonsensical)}. This work-in-progress currently \emph{focuses on memorization}. We propose a novel system that can determine the amount of potential memorization in model-generated recipes. The system is based on the overlap between (i) the generated recipe, considering ingredients and recipe steps, and (ii) retrieved online documents on which it could have been trained. The pipeline consists of (i) generating a recipe using an LLM, (ii) analysing the recipe for non-sensicality and general correctness\footnote{Not containing any significant mistakes or flaws.}, and (iii) performing the overlap examination described above. 

To tune and test our system, we first let human annotators follow the steps of our system and then verify which LLMs to use in different steps of our system in order to close the gap to humans.

In summary, this work-in-progress presents:

\begin{enumerate}
    \item An annotation approach that quantifies amounts of memorized and creatively generated ingredients and recipe steps that both humans and LLMs can follow.
    \item A dataset of nearly $36,000$ instances ($20$ recipes, each accompanied with annotation of ingredients and task components from $18$ documents) of human annotations that allow us to (a) propose answers to our research questions, (b) tune and automate an LLM-as-judge system that replicates the annotation process and (c) estimate the gap between humans and best LLM-as-judge systems.
    % \item A system capable of detecting and quantifying memorization and creativity in recipes generated by LLMs.
    \item Our work-in-progress and future plan on automating the proposed annotation approach with LLMs, which would allow us to scale up our study beyond $20$ recipes and a single LLM recipe generator.
\end{enumerate}

%We consider a task or ingredient to be creatively generated if this task or ingredient can not be found in online sources, while still being acceptable in the context of the recipe. A task or ingredient that is not acceptable in the context of the recipe (e.g. being incorrect, dangerous or otherwise not working) is deemed as non-sensical.

\section{Related Work}

RecipeMC~\cite{taneja2024montecarlotreesearch}, a text generation method for recipes, uses a fine-tuned GPT-2~\cite{radford2019language} and Monte Carlo Tree Search for recipe ingredient and steps generation. Other prompt-based recipe generation methods were proposed \cite{H_Lee_2020, hwang2023largelanguagemodelssous}. These approaches differ from ours, as we use models that are pre-trained but not fine-tuned for recipe generation. Our prompting approach also differs.
%Helena \citet{H_Lee_2020} present RecipeGPT, a recipe generation and evaluation system utilizing GPT-2, fine-tuned on a large cooking recipe dataset. 
%\citet{hwang2023largelanguagemodelssous} present their work on prompt engineering and development in the cooking recipe area in their paper.

\citet{chen-etal-2024-multi-perspective} research memorization by collecting model outputs (tokens, probability distributions, embeddings) and calculate memorization score for every sentence. This differs from our approach, as we evaluate memorization based on tags (ingredients, tasks, tools and ingredients used in the tasks).
\citet{hartmann2023sokmemorizationgeneralpurposelarge} focus on memorization, the challenges with its definition and estimation, and identifies various types of information an LLM can be said to memorize (facts, verbatim text, algorithms, etc.), as well as the potential related risks and measurement of LLM memorization, its impact, and its mitigation. A mentioned approach to estimating memorization in ideas and algorithms is prompting a model with all but a single element or step (tuple completion), and checking whether the true missing element or step is predicted by the model. 

\citet{lee_plagiarization} define plagiarism as reusing of any content including text, source code, or audio-visual content without the permission or citation from the author of the original work. They define 3 types of plagiarism: \emph{verbatim}, \emph{paraphrase} and \emph{idea} plagiarism. Our definition of \emph{potential plagiarism} is more specific. We attempt to detect all three mentioned types with our approach.

\citet{lee2025plagbenchexploringdualitylarge} introduce PlagBench, a dataset that represents various types of plagiarism (verbatim copying, paraphrasing, summarization), and they utilise this dataset to examine LLM's ability to transform original content and evaluate plagiarism detection performance of LLMs.

\citet{stochastic_parrots} explore how LLMs, often trained using large, undocumented datasets, often containing unchecked information, can cause hallucination. They use the term ``stochastic parrot''---the model does not understand the language, it simply ``parrots'' training data and statistical patterns found in the dataset~\cite{stochastic_parrots_webpage}.

%\citet{gui2024bonbonalignmentlargelanguage} focus on aligning samples from LLMs to human preferences using best-of-$n$ sampling.
%\citet{pezeshkpour-hruschka-2024-large} examine positional sensitivity in multiple-choice questions and selections, and work to calibrating LLM's predictions to improve their capabilities.

%\citet{zhang2024confusedcautiouslytextualsequence} propose Entropy Maximization with Selective Optimization, a framework that focuses on Textual Sequence Memorization erasure for LLMs with a focus on a good erasure-utility trade-off. 

%\citet{carlini_memorization} focus on memorization and the extraction of training data from LLMs, proposing a way to extract memorized and possibly private data and strategies to mitigate privacy leakages from LLMs. 
%\citet{Zhao2024AssessingAU} work on assessing creativity in LLMs, designing a framework to automatically assess creativity and adapting and modifying the Torrance Tests of Creative Thinking~\cite{torrance1966torrance}.
%\citet{Runco2012TheSD} define creativity and debate the necessity of the criteria for creativity (originality, effectiveness). They overview literature focusing on effectiveness and creativity as a whole. 
%\citet{Luo2024HallucinationDA} focus on hallucination detection and mitigation. They cover various classification metrics, archetypes of hallucination detection, and methods of hallucination mitigation.
%\todo{Add related work, definite creativity, halucinací, memorizace}

\section{Research Process Overview}
\label{section:research_process}

For brevity, we define the following terms:

\begin{description}[style=unboxed,leftmargin=0em,listparindent=\parindent]
     \setlength\parskip{0em}
    \item[Recipe name] The name of a specific dish (e.g. ``Coleslaw'', ``Beef wellington'').
    \item[Ingredient] A single item (e.g. spice, food, liquid) that, combined with other items (ingredients) in a specific way, forms a dish (e.g. bell pepper, whole milk, salt).
    % \item[Task] A single action that is being performed (e.g. ``Preheat an oven.'').
    \item[Task] A triple consisting of (i) the task name and optionally (ii) the tools involved and/or (iii) the ingredients involved, characterizing a single performed action (e.g. ``Preheat an oven'', ``Mix flour and salt in a large mixing bowl''). 
    \item[Step] A single task or a combination of tasks (e.g. ``First, we need to preheat the oven and mix together the dry ingredients'').
    \item[Recipe] A triple consisting of a recipe name, a list of ingredients and a list of steps.
\end{description}

To analyze to what extent the LLMs generate creative, memorized, or non-sensical recipes, we devise the following research process:

% \begin{enumerate}
%     \item Proposal of the list of candidate recipe names. Selection of recipe names for further examination (Section~\ref{RecipeGeneration}).
%     \item For each recipe name, do N-way recipe generation (Section~\ref{RecipeGeneration}). 
%     \item Selecting best-of-N recipe, non-sense detection, filtering out non-sense and wrongly generated recipes (Section~\ref{RecipeGeneration}).
%     \item Parsing of model-generated recipes for annotation (Section~\ref{Parsing}).
%     \item Retrieval of online documents for each recipe (Section~\ref{AnnotationSystem}).
%     \item Annotation of online documents (Section~\ref{AnnotationSystem},~\ref{AutomatizedAnnotation}).
%     \item Gathering of statistical data from the annotation, interpretation of the data (Section\ref{ResultsHumans}) \todo{Přidat výsledky strojů}.
% \end{enumerate}

\begin{description}[style=unboxed,leftmargin=0em,listparindent=\parindent]
     \setlength\parskip{0em}
    \item[(1) Candidate Recipe Names Selection] Proposal of the list of candidate recipe names. Selection of recipe names for further examination (Section~\ref{RecipeGeneration})
    \item[(2) LLM Recipe Generation] For each recipe name, do $K$-way recipe generation (Section~\ref{RecipeGeneration}). 
    \item[(3) Filtering] Selecting best-of-$K$ recipe, detection of objectively wrong steps and ingredients, filtering out non-sense and wrongly generated recipes (Section~\ref{RecipeGeneration}). Examples can be found in Appendix~\ref{appendix:nonsense_examples}. 
    \item[(4) Parsing] Parsing of model-generated recipes for annotation (Section~\ref{AnnotationSystem}).
    \item[(5) Online Document Retrieval] Retrieval of online documents for each recipe (Section~\ref{AnnotationSystem}).
    \item[(6) Annotation] Annotation of the online documents (Section~\ref{AnnotationSystem},~\ref{AutomatizedAnnotation}).
    \item[(7) Gathering Statistics] Gathering of statistical data from the annotation and interpretation of the data (Section~\ref{ResultsHumans}). 
\end{description}

Our system seeks to automate steps $2$ to $6$. This approach is different from existing approaches, such as that of~\citet{hartmann2023sokmemorizationgeneralpurposelarge}, who suggest prompting a model with all but a single element or step, and detecting whether the true missing element is generated by the model. 

\section{Recipe Generation}
\label{RecipeGeneration}

With the purpose of creating the annotation dataset, we manually created a list of $62$ recipe names from around the world \textbf{(step~(1))} while focusing on (i) variety ($23$~countries/regions) and (ii) not targeting only popular recipes (such as national dishes of major countries). Out of these, we selected $20$ recipes for closer examination (list available in Appendix~\ref{appendix:RecipeNames}). 

%We used ChatGPT~\cite{chatgpt} due to the ease of prompting and casual experimenting. 
We devised $5$ prompt types with varying levels of detail to try to select recipes to select the prompt that returns the best results (described in more detail in Appendix~\ref{PromptTypes})\footnote{Automated pipeline will offer the ability to use custom prompts.}. The prompt selection was performed by applying the previously described prompt styles in ChatGPT-3.5 and Mixtral and evaluating the outputs, utilising SPA~\cite{ethayarajh-jurafsky-2022-authenticity} in order to select the prompt type resulting in the most satisfactory results. Using this prompt type, we performed generation with $K = 5$, using $5$ distinct prompts of the same type to obtain $5$ generated recipes.
%The ``More detail'' prompt type (e.g. ``Give me a detailed guide to making fish and chips'') was the most preferred. 
%Using Mixtral~\cite{jiang2024mixtralexperts}, we then performed 5 generations per recipe using the ``More detail'' prompt type \textbf{(step~(2))}.

%
%The prompts containing a request for greater detail yielded the best subjective results.

%We engineered the prompts and generated the output of both ChatGPT and Mixtral for 5 recipes, and using the SPA ranking method (comparing each output with each output and distributing 100 points among them), we selected the prompt templates most preferred by a human. While the results were inconclusive for ChatGPT (version 3.5 at the time of research), Mixtral 8x7B Instruct v0.1~\cite{jiang2024mixtralexperts} resulted in the best SPA ranking for the ``More detail'' prompt style. Thus, for further prompting, we selected the ``More detailed'' prompt style, with the templates being the winning prompts for each of the 5 recipes (such as ``Give me a detailed walkthrough to making [recipe name]'' or ``Give me a detailed guide to making [recipe name]'').

\section{Annotation System}
\label{AnnotationSystem}

Following the generation, we select a single best-of-$K$ recipe, based on (i) the amount of objectively faulty instructions or ingredients (e.g. using a paper towel to extract moisture from grated cucumber---the paper towel would fall apart), (ii) a subjective opinion of correctness and  (iii) the general preference of the generated output \textbf{(step~(3))}, following the stated order. 

Once recipes are generated and filtered, they need to be parsed \textbf{(step~(4))}. The parsing of ingredients is done by parsing the provided ingredient list. To parse tasks, we form triples as defined in Section~\ref{section:research_process}.
%, we define every task in a recipe as a triple composed of (i) the task name, (ii) the tools involved in the task, and (iii) the ingredients involved in the task. 
The task name was usually selected from a verb, sometimes including the preposition (e.g. ``whisk'', ``pour in''). It was parsed in the infinitive form. The tool or the ingredient was occasionally not present in a step and was thus not parsed. 
%(for example, in the step ``Heat a large frying pan on high heat'', an ingredient is not explicitly mentioned or involved).
The tool involved in a task was parsed and propagated into the following relevant tasks, even if it is not explicitly mentioned. 
The list of ingredients was parsed the same way. In the case of an implied ingredient or multiple ingredients, a word was selected to sum all of the ingredients (e.g. ``mixture'').

%A total of 18 documents are retrieved for each of the 20 recipes \textbf{(step~(5))}\footnote{In automated pipeline, higher tens of documents will be retrieved using APIs and Python libraries to broaden the dataset.}. 

A total of $N_d$ documents are retrieved for each examined recipe. These documents are obtained in the form of websites containing a recipe, and are then documents are then annotated.

The ingredient annotation \textbf{(step~(6))} is based on locating each model-generated ingredient verbatim in a parsed document's ingredient list. The annotator has 4 options during the annotation: \emph{``Found''}, \emph{``Found (not perfect)''}, \emph{``Not found''} and \emph{``Implied''}\footnote{Option merged with ``Not found'' in the automatized annotation.}. 

For each task, there are $3$ different annotation fields the annotator has to select an option in: (1) the task, (2) the tool(s), and (3) the ingredient list. 
For the tasks, the annotator selects one of four options: \emph{``Task Found''}, \emph{``Task Found (Not Exact Wording)''}, \emph{``Task Found (Wrong Context)''}, and \emph{``Task Not Found''}.

For the tools, the annotator selects one of $5$ options: \emph{``Found''}, \emph{``Found (Not Exact)''}, \emph{``Not Found''}, \emph{``Tool Implied''}, and \emph{``No Tool Involved''}. 

For ingredients in the context of tasks, the selection is slightly different due to the involvement of multiple ingredients in a single task. For that reason, we focus on the presence of the ingredients regardless of whether they match the wording in the document verbatim. The options for the annotator to choose from are: \emph{``Ingredients Match''}, \emph{``Most Ingredients Match''}, \emph{``Some Ingredients Match''}, \emph{``No Ingredients Match''}, \emph{``Ingredients Implied''}, and \emph{``No Ingredients Used''}.

For an in-depth overview of the annotation system, see Appendix~\ref{appendix:AnnotationGuide}.

\section{Human Annotation and  Findings}
\label{ResultsHumans}

For the purposes of human annotation, we obtained $N_d = 18$ for $20$ individual recipes for a total of $360$ documents. These documents were selected manually, with the requirement that the document must contain a list of ingredients and a list of steps. $6$ documents were retrieved from $3$ search engines (Google\footnote{\url{https://www.google.com/}}, Bing\footnote{\url{https://www.bing.com/}} and DuckDuckGo\footnote{\url{https://duckduckgo.com/}}). We used Mixtral~\cite{jiang2024mixtralexperts} to perform the $K = 5$ generations.

We utilised the help of $7$ annotators (undergraduate and graduate students, one of whom was the first author), who took $148$ hours to annotate the dataset ($7.4$ hours per recipe, or $16$ minutes for a single document annotation). The annotators were paid $175$ CZK per hour for their work. 
Half ($180$) of the documents were annotated twice, with each annotator having an overlap of 6 out of the 9 annotated documents with other annotators. This choice was made to be able to calculate inter-agreement.

With the described system, we gathered %nearly 36,000 data points 
$6669$ annotation instances for ingredients and $9693$ task triples annotation instances for the $18 \times 20$ documents.  The collected dataset provides (i) information necessary to tune an automatic annotation system and evaluate the capabilities of various LLMs and (ii) allows us to extract preliminary results targeting our research questions \textbf{(step 7))}. The annotators followed a prepared annotation guide (can be found in~\ref{appendix:AnnotationGuide}).

\subsection{Ingredients Annotation}
\label{HumanIngAnnotation}

We calculated the inter-agreement using Cohen's~$\kappa$~\cite{Cohens_Kappa}, with the macro average value\footnote{Macro average value was calculated as a sum of $\kappa$ score for each recipe-annotator pair divided by the number of recipe-annotator pair doubles ($60$).} being $0.77$. To establish human performance, the macro accuracy between annotators $A_h$ is defined as:
$$A_h = \frac{\sum_{j=1}^{m} \frac{\sum_{i=1}^{n} \frac{S_{ij}}{T_{ij}}}{n}}{m}$$ 
where $n$ is the number of annotated recipes, $m$ is the number of annotator pairs (e.g. for $3$ annotators, there are $3$ annotator pairs ($1,2;  1,3; 2,3$), $S_{ij}$ is the number of annotations from recipe $i$ where two annotators from the pair $j$ chose the same option, and $T_{i,j}$ is the total number of annotations in recipe $i$ the two annotators from the pair $j$ shared. In our annotation, this macro accuracy was observed to be 85.23\%.

%We found out that while most of the observed inter-agreements were satisfactory at 0.7 or higher, there were occasional flukes. The lowest observed Cohens Kappa value were 0.2, 0.48 and 0.51. However, the average Cohens Kappa across all recipes and all annotator combinations was 76.78\%.

%\begin{table}[h]
%\centering
%\begin{tabular}{|l|r|r|r|}
%\hline
%Annotator Pair & Avg. & Max. & Min.\\
%\hline
%All Annotators & 0.77  & 1.00 & 0.20\\
%Ann1 \& Ann2 & 0.74  & 0.96 & 0.49\\
%Ann1 \& Ann3 & 0.77  & 1.00 & 0.20\\
%Ann2 \& Ann3 & 0.79 & 1.00 & 0.48\\
%\hline
%\end{tabular}
%\caption{Cohen's Kappa for Annotator Pairs in Ingredient Annotation.}
%\label{tab:kappa}
%\end{table}

%\todo{Tabulka: Sloupce: Highest, Lowest, Avg, Median}

When considering all annotations of all recipes in all documents, the most selected option was ``Found''. ``Implied'' was the rarest choice. The summary with details is available in Table~\ref{tab:ing_table}.

\begin{table}[ht!]
\centering
\begin{tabular}{|l|r|r|}
\hline
Selection & Count & Percentage \\
\hline
Found & 2569 & 38.82\,\% \\
Found (not perfect) & 1979 & 29.67\,\% \\
Not found & 1996 & 29.93\,\% \\
Implied & 105 & 1.57\,\% \\
\hline
Total & 6669 & 100.00\,\%\\
\hline
\end{tabular}
\caption{Ingredient Annotation Selection Summary. Percentage rounded to 2 decimal places.}
\label{tab:ing_table}
\end{table}

%The key part of the analysis focused on finding ingredients that the model has generated, but they were not found perfectly by any of our annotators in any of the documents. Our analysis showed only 8 out of the total 247 such ingredients in the dataset (shown in Table~\ref{tab:NotFoundIngs}).

$8$ of the $247$ observed ingredients were not found perfectly by the annotators (Table~\ref{tab:NotFoundIngs}).

\begin{table}[h]
\begin{adjustbox}{width=\columnwidth,center}
\begin{tabular}{ |l|l||l|l|  }
 \hline
 Recipe               & Ingredient Name      & FNP & NF \\
 \hline
 Paella               & Yellow Bell Pepper   & $3$   & $24$  \\
 Leberkäse            & Allspice             & $0$   & $27$  \\
 Chilli con carne     & Black Beans          & $0$   & $27$  \\
 Chocolate mousse     & Whole milk           & $0$   & $27$  \\
 Coleslaw             & Mustard              & $14$  & $13$  \\
 Thai Green Curry     & White pepper         & $8$   & $19$  \\
 Koshari              & Green lentils        & $26$  & $1$   \\
 Thai Green Curry     & Snap peas            & $6$   & $21$  \\
 \hline
\end{tabular}

\end{adjustbox}
\caption{Ingredients not found perfectly by annotators. FNP is number of ``Found (not perfect)'' selections, NF is number of ``Not found'' selections. The detailed explanation is in annotation guidelines (Appendix~\ref{sub:ing_annotation_guide}).}
\label{tab:NotFoundIngs}
\end{table}

We hypothesize that the missing ingredients either (i) \emph{hint that model made a creative action by introducing a new ingredient} or (ii) \emph{the artifact was only produced by the non-exhaustivity of our annotation}---annotators ``only'' checked $18$ documents for each recipe. To validate these hypotheses, we manually repeated the search query within the search engines, while this time appending the name of the not found ingredient. In case of every missing ingredient, we managed to find a relevant document with recipe containing such an ingredient. Therefore, based on human annotation, \textbf{we conjecture that hypothesis (ii) is correct, and Mixtral model uses ingredients only and only if they were seen in the training set}.

\subsection{Tasks Annotation}

%Compared to Ingredients Annotation, 
We observed $\kappa$ to be lower than in the case of the Ingredients Annotation, with the macro average value\footnote{Macro average value was calculated the same way as in ingredient annotation, with task name, tool and ingredient annotation contributing equally.} being $0.63$. The observed macro accuracy was $74.15\,\%$.

%\begin{table}[h]
%\centering
%\begin{tabular}{|l|r|r|r|}
%\hline
%Annotator Pair & Avg. & Max. & Min.\\
%\hline
%All Annotators & 0.63  & 0.96 & 0.35\\
%Ann1 \& Ann2 & 0.63  & 0.84 & 0.35\\
%Ann1 \& Ann3 & 0.63  & 0.93 & 0.36\\
%Ann2 \& Ann3 & 0.63 & 0.96 & 0.35\\
%\hline
%\end{tabular}
%\caption{Cohen's Kappa for Annotator Pairs in Task Annotation.}
%\label{tab:kappa_tasks}
%\end{table}

%The analysis of annotator selections for tasks~(Table~\ref{tab:table_tasks} in Appendix~\ref{appendix:AnnotationSummary}) showed similarities to how the annotators chose in the ingredient annotation section. A significant portion of the tasks was marked as ``Not Found''.

\begin{description}[style=unboxed,leftmargin=0em,listparindent=\parindent]
     \setlength\parskip{0em}
    \item[Task Annotation] The task annotation showed that the percentage of ``Task Not Found'' selections was the highest, followed by the number of selections of ``Task Not Found'' and ``Task Found (Not Exact Wording)''.
    \item[Tool Annotation] Influenced by the number of ``Task Not Found'' selections, the tool annotation had nearly half of the values not filled in. Of those filled in, ``Found (Not Exact Wording)'' was the most selected option, followed by ``Tool Implied'' and ``Not Found'', with ``Found'' being the least selected. 
    \item[Ingredient List Annotation] Similar to the task annotation, the ingredient list annotation had nearly half of the values not filled in, and the number of ``Ingredients Match'' selections greatly outnumbered the other selections, with the other $4$ categories only totalling $16.94\,\%$ of the selections.
\end{description} 

The summary with all the details can be found in Tables~\ref{tab:table_tasks},~\ref{tab:table_tasks_tools} and~\ref{tab:table_tasks_ingredients}.

\begin{table}[ht!]
\centering
\begin{tabular}{|p{3.7cm}|r|r|}
\hline
Selection & Count & Percentage \\
\hline
Task Found & 3259 & 33.62\,\% \\
Task Found (Not Exact Wording) & 2402 & 24.78\% \\
Task Found (Wrong Context) & 591 & 6.10\,\% \\
Task Not Found & 3441 & 35.50\,\% \\
\hline
Total & 9693 & 100.00\,\%\\
\hline
\end{tabular}
\caption{Task Annotation Selection Summary for Tasks. Percentage rounded to 2 decimal places.}
\label{tab:table_tasks}
\end{table}

\begin{table}[ht!]
\centering
\begin{tabular}{|p{3.7cm}|r|r|}
\hline
Selection & Count & Percentage \\
\hline
Found & 788 & 8.13\,\% \\
Found (Not Exact Wording) & 1979 & 20.42\,\% \\
Not Found & 792 & 8.17\,\% \\
Tool Implied & 1601 & 16.52\,\% \\
\textit{Not filled in} & 4533 & 46.76\,\%\\ 
\hline
Total & 9693 & 100.00\,\%\\
\hline
\end{tabular}
\caption{Task Annotation Selection Summary for Tools. Percentage rounded to 2 decimal places. The value was often not filled in due to the annotation system (if the task was not found, the tool checkbox was left empty, or the lack of a tool in the step).}
\label{tab:table_tasks_tools}
\end{table}

\begin{table}[ht!]
\centering
\begin{tabular}{|l|r|r|}
\hline
Selection & Count & Percentage \\
\hline
Ingredients Match & 3603 & 37.17\,\% \\
Most Ingredients Match & 551 & 5.68\,\% \\
Some Ingredients Match & 232 & 2.39\,\% \\
No Ingredients Match & 252 & 2.60\,\% \\
Ingredients Implied & 607 & 6.26\,\% \\
\textit{Not filled in} & 4448 & 45.89\,\%\\ 
\hline
Total & 9693 & 100.00\,\%\\
\hline
\end{tabular}
\caption{Task Annotation Selection Summary for Ingredients. Percentage rounded to 2 decimal places. The value was often not filled in due to the annotation system---if the task was not found, the ingredient list checkbox was left empty, or the lack of an ingredient list in the step.}
\label{tab:table_tasks_ingredients}
\end{table}

\FloatBarrier

$29$ ($8.1\,\%$) tasks were never marked as ``Task Found''. These tasks can be found in Table~\ref{tab:TasksNotFoundPerfectly}. \textbf{This hints at a considerable amount of potential creativity when it comes to Tasks in recipes.} We have yet to analyze whether the lack of ``Task Found'' selections is caused by a lack of exhaustivity---suggesting Mixtral is also a task plagiator. A similar analysis is needed for ingredient lists and tools, too.

% Post table from RMLLM GitHub Repository/Annotation_examination/stats_tasks.ipynb
\begin{table}[h]
\begin{adjustbox}{width=\columnwidth,center}
\centering
\begin{tabular}{|l|l||l|l|l|}
\hline
Recipe & Task & TFNEW & TNF & TFWC \\
\hline
Chocolate Mousse & Stir into & $0$ & $24$ & $3$ \\
Fish and Chips & Rinse & $0$ & $27$ & $0$ \\
Fish and Chips & Increase oil temp & $11$ & $15$ & $1$ \\
Gyoza & Add & $0$ & $9$ & $18$ \\
Gyoza & Turn onto & $1$ & $26$ & $0$ \\
Gyoza & Cut out & $1$ & $26$ & $0$ \\
Gyoza & Place into & $25$ & $2$ & $0$ \\
Gyoza & Pour in & $27$ & $0$ & $0$ \\
Koshari & Ladle & $20$ & $7$ & $0$ \\
Lagman & Cut into strips & $7$ & $19$ & $1$ \\
Lagman & Gently drop & $9$ & $18$ & $0$ \\
Leberkäse & Whisk together & $1$ & $26$ & $0$ \\
Leberkäse & Press down & $7$ & $20$ & $0$ \\
Leberkäse & Remove from oven & $5$ & $22$ & $0$ \\
Pastel da nata & Press into the wells & $25$ & $2$ & $0$ \\
Pita Bread & Place into & $13$ & $13$ & $0$ \\
Tzatziki & Press on & $17$ & $10$ & $0$ \\
\hline
\end{tabular}
\end{adjustbox}
\caption{List of tasks that were not found perfectly during the annotation by any human annotator. $12$ tasks from the recipe ``Fettuccine Alfredo'' were omitted from the table because all examined documents used premade pasta while the model provided steps for making pasta from scratch. TFNEW is number of ``Task Found (Not Exact Wording)'' selections, TNF is number of ``Task Not Found'' selections and TFWC is number of ``Task Found (Wrong Context)'' selections. For closer explanation, refer to Appendices ~\ref{subsub:tasks_start}-\ref{subsub:task_last}.}
\label{tab:TasksNotFoundPerfectly}
\end{table}

%As the tool annotation results show (Table~\ref{tab:table_tasks_tools} in Appendix~\ref{appendix:AnnotationSummary}), we can see the results of nearly 35.5\% of the tasks being annotated as ``Not Found''. Nearly 47\% of all selections were left empty. Of the remaining options, the most prevalent one was ``Found (Not Exact Wording)'', followed by ``Tool Implied''. This was expected, as in many cases, the instructions in a document only mention the tool once at the very start. During parsing, however, the tools were propagated further.

%We can observe a similar percentage of annotation objects not being annotated in the ingredient list annotations (Table~\ref{tab:table_tasks_ingredients} in Appendix~\ref{appendix:AnnotationSummary}). This is due to either a task being annotated as ``Not Found'', or the ingredient list not being present in the parsed task. We can, however, observe a high amount of perfectly matching ingredient lists. The number of matching ingredient lists is substantially higher than the number of all the other selections.

% Figures~\ref{fig:percentages} and~\ref{fig:percentages_tasks} show how combining the annotated documents can result in very high numbers of ``Found'' selections in ingredient annotation and ``Task Found'' selections in tasks annotation, respectivelly. The growing trend makes us confident that with more documents being covered, the higher the chance is that we find ingredients and tasks not found in already examined documents.

\subsection{Exhaustivity Analysis}

\begin{figure}[ht!]
    \centering
    \includegraphics[width=0.48\textwidth]{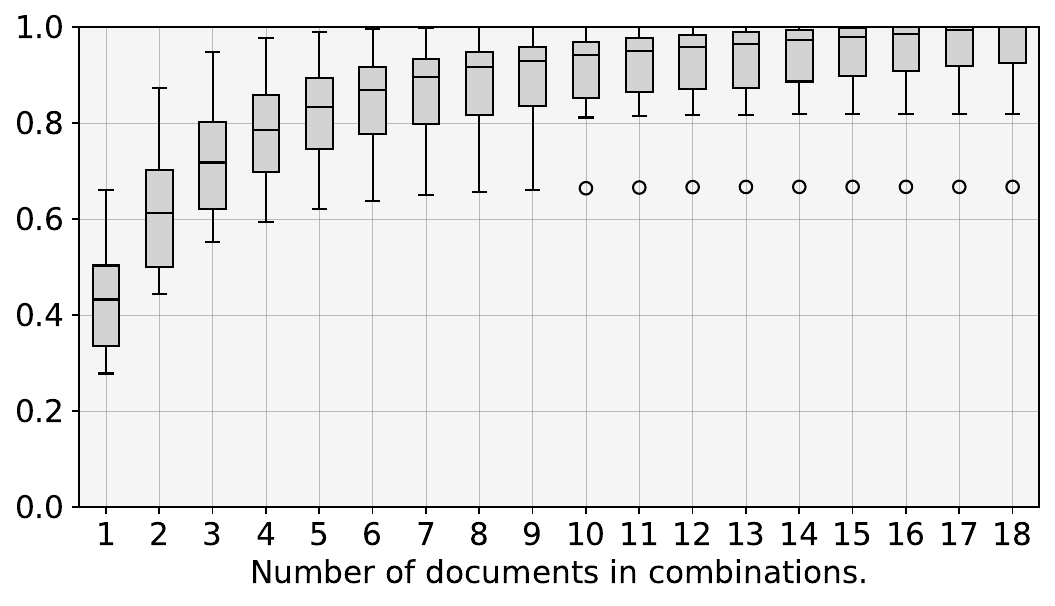}
    \caption{Percentage of Found selections in a combination of $n$ documents in ingredients annotation across all recipes. The combination of higher amounts of documents shows that the more documents are examined, the more ingredients are annotated as ``Found''.}
    \label{fig:percentages}
\end{figure}

\begin{figure}[ht!]
    \centering
    \includegraphics[width=0.48\textwidth]{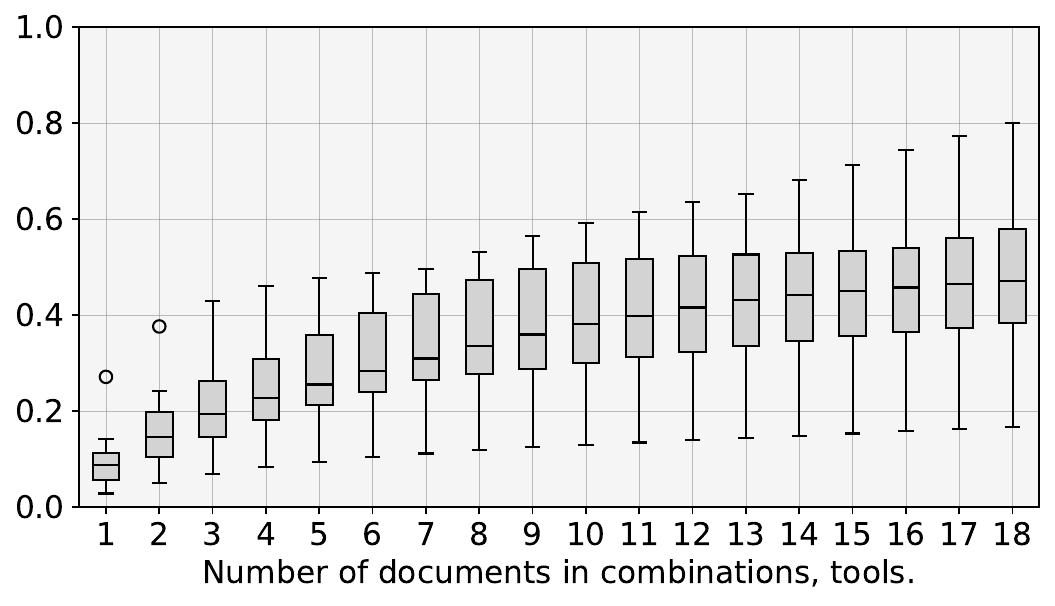}
    \caption{Percentage of ``Found'' selections in a combination of $n$ documents in tools annotation across all recipes. We can observe the rising level of coverage, hinting at the need for more examined documents.}
    \label{fig:percentages_tools_tasks}
    %\vspace{-0.5cm}
\end{figure}

\begin{figure}[ht!]
    \centering
    \includegraphics[width=0.48\textwidth]{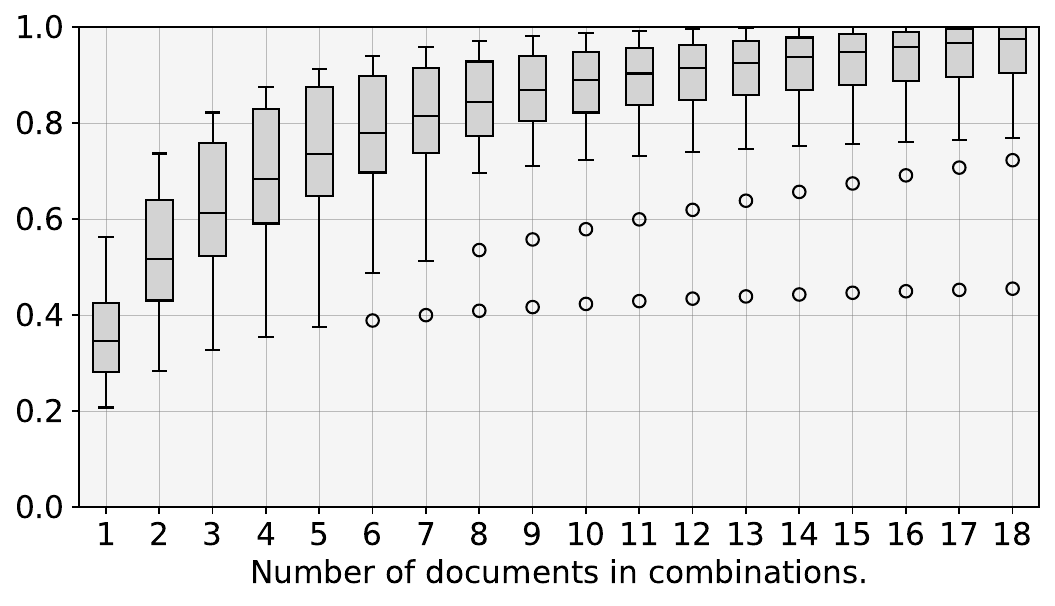}
    \caption{Percentage of ``Task Found'' selections in a combination of $n$ documents in tasks annotation across all recipes. The combination of higher amounts of documents shows that the more documents are examined, the more tasks are annotated as ``Task Found'', but the growth is slower compared to ingredients.}
    \label{fig:percentages_tasks}
    %\vspace{-0.5cm}
\end{figure}

\begin{figure}[ht!]
    \centering
    \includegraphics[width=0.48\textwidth]{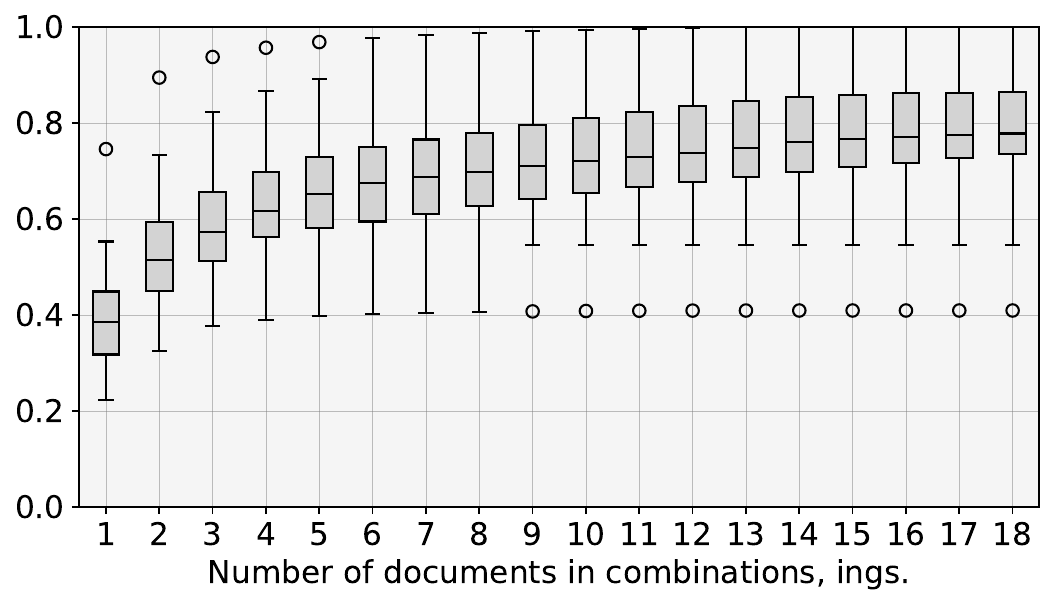}
    \caption{Percentage of ``Ingredients Match'' selections in a combination of $n$ documents in ingredients annotation in tasks across all recipes.}
    \label{fig:percentages_ingredients_tasks}
\end{figure}

Finally, we analyze the exhaustivity of annotation on $N_d = 18$ documents. The more documents are covered, the higher the chance that full coverage is achieved and all ingredients/tasks in a recipe generated from an LLM are found to be contained in a document or their combination. We observe that for more challenging tasks such as tool item annotation, \emph{the proportion of found items in Figure~\ref{fig:percentages_tools_tasks} saturates more slowly} compared to ingredient annotation in Figure~\ref{fig:percentages} (additionally, Figures \ref{fig:percentages_tasks} and \ref{fig:percentages_ingredients_tasks} show how ``Task Found'' and ``Ingredients Match'' selections saturate the more documents are combined). 

As we demonstrate by these plots, automatized annotation will allow us to cover a broader range of documents ($N_d = 80$ planned for automatized annotation), reducing the problems associated with non-exhaustivity of human annotations. We observed that in ingredient annotation, just $5$ documents need to be combined to achieve a proportion of found ingredients of near $100\,\%$.

\section{Automatized Annotation}
\label{AutomatizedAnnotation}

%\includegraphics[width=0.5\textwidth]{Fig/automatized_diagram.pdf}
%\captionof{figure}{Automatized Annotation Pipeline. 357 documents were obtained. 96.7\% to 98.9\% extracted successfully depending on the LLM.}

%Out of the 360 retrieved documents that were annotated, 
The initial part of the automatized annotation is the assessment of models and their ability to annotate the same way human annotators would. A total of $357$ out of $360$ documents were still available when the automatized annotation part of the research started. We downloaded these documents in the HTML format for further processing.

We processed the retrieved documents with $3$ LLMs (Mixtral 8x7B Instruct v0.1, Mixtral 7B Instruct v0.3~\cite{jiang2023mistral7b} and Llama 3.1 8B Instruct~\cite{llama}). First, we instructed LLMs to extract the list of ingredients into an XML structure. 

Secondly, an annotation prompt is formed, containing (i) the name of the annotated ingredient, task, tool, or ingredient list related to the task, (ii) the extracted list of items (again ingredients, tasks, tools, or ingredient lists) from the generated recipe, and (iii) the task description. 

The automated annotation utilises the Cloze Formulation (CF) approach defined in the Open Language Model Evaluation Standard (OLMES)~\cite{olmes}. 
The model is presented with N prompts (where N is the number of choices in the annotation type) that differ only in the final choice. The LLM probability for the chosen tokens is then used to predict the answer of the model. We were then able to calculate the accuracy of various models working with the extracted lists using the following formula: 

$$A_m = \frac{\sum_{i = 1}^{m} \frac{S_i}{T_i}}{m}$$

Where $m$ is the number of recipes, $S_i$ is the number of document-ingredient/task/tool/ingredient list pairs that the model annotated the same way as the human annotator in recipe $i$, and $T_i$ is the total number of pairs in recipe $i$. For the purposes of the calculation, in the case of a document being annotated twice (as described in Section~\ref{ResultsHumans}), a pair is counted towards $S_i$ in the case that the model annotated the same way as either of the two human annotators. If a document was annotated twice, each pair is only counted once towards $T_i$.

The ability of the models to annotate ingredients the same way as humans is summarized in Table~\ref{tab:accuracy_models}. We observed the best accuracy from Gemma 2 9B Instruct~\cite{gemmateam2024gemma2improvingopen} and a very problematic performance by Llama 3.1 8B Instruct~\cite{llama}.

\begin{table}[h]
\centering
\begin{adjustbox}{width=\columnwidth,center}
\begin{tabular}{|l|l||P{1.8cm}|}
\hline
Annotation Model &  Extraction Model & Accuracy \\
\hline
\hline
 & Mixtral 8x7B Instruct v0.1 & $0.631$ \\
Mixtral 8x7B Instruct v0.1 & Mistral 7B Instruct v0.3 & $0.674$ \\
 & Llama 3.1 8B Instruct & $0.708$ \\
\hline
 & Mixtral 8x7B Instruct v0.1 & $0.592$ \\
Mixtral 7B Instruct v0.3 & Mistral 7B Instruct v0.3 & $0.602$ \\
 & Llama 3.1 8B Instruct & $0.589$ \\
\hline
 & Mixtral 8x7B Instruct v0.1 & $0.707$ \\
Gemma 2 9B Instruct & Mistral 7B Instruct v0.3 & $0.740$ \\
 & Llama 3.1 8B Instruct & \textbf{$0.778$} \\
\hline
 & Mixtral 8x7B Instruct v0.1 & $0.377$ \\
Llama 3.1 8B Instruct & Mistral 7B Instruct v0.3 & $0.369$ \\
 & Llama 3.1 8B Instruct & $0.355$ \\
\hline
 & Mixtral 8x7B Instruct v0.1 & $0.685$ \\
Gemma 2 27B Instruct & Mistral 7B Instruct v0.3 & $0.716$ \\
 & Llama 3.1 8B Instruct & $0.745$ \\
\hline
\end{tabular}
\end{adjustbox}
\caption{Model Statistics for $3$ classes (Found, Found (not perfect) and Not found) in the ingredients annotation. Using human annotation as ground truth. Formula used to calculate accuracy can be found above.}
\label{tab:accuracy_models}
\end{table}

This step of the research was done to assess various models in order to select the best model to use for the final stages of the research when the entire pipeline is fully automatized. It is notable that the extraction by Llama 3.1 8B Instruct resulted in the highest accuracy in annotations with $3$ out of the $5$ tested models. 

The assessment of models for task annotation proved to be more complicated, and showed much worse results. The tasks, together with tools and ingredient lists, were extracted into an XML structure. When $4$ classes were considered during the task name annotation, the examined models showed an accuracy of only $35-50\,\%$ depending on the model. After tuning the prompt used for the extraction of task lists from documents and reducing the number of annotation classes to 2, we observed an accuracy of close to $75\,\%$ (as can be seen in Table~\ref{tab:accuracy_models_tasks}). In the future, we aim to evaluate the precision-recall curve, setting a precision value and reporting recall for the set precision.

\begin{table}[h]
\centering

\begin{adjustbox}{width=\columnwidth,center}
\begin{minipage}{\columnwidth}
\centering
\small
\begin{tabular}{|l|P{1.8cm}|}
\hline
Annotation Model & Accuracy \\
\hline
\hline
Mixtral 8x7B Instruct v0.1 & $0.74$ \\
Gemma 2 9B Instruct & $0.75$ \\
Llama 3.1 8B Instruct & $0.71$ \\
Mistral 7B Instruct v0.3 & $0.71$ \\
\hline
\end{tabular}
\end{minipage}
\end{adjustbox}
\caption{Model Statistics for $2$ classes in task name annotation. Identical extracted dataset (extraction performed using the Llama 3.1 8B Instruct model) used in all annotations.}
\label{tab:accuracy_models_tasks}
\end{table}

\section{Ongoing and Future Work}

Our findings so far are based on human annotation and assessment of models based on human annotations. The current ongoing step of the research is assessing models and their performance when following the same annotation steps as humans. While the annotation system and the assessment of models for the annotation of ingredients were successfully completed, the assessment of models for the annotation of tasks posed unique challenges that are yet to be fully overcome.

Our next steps are to (i) complete the assessment of models in automatized task annotation, 
%(ii) automatize steps $2$ through $6$ of the research process (Section \ref{section:research_process}) using models that were observed to perform the best for annotation, 
(ii) complete the automatization of steps $2$ through $6$ of the research process (steps $2$ and $4$ utilising LLMs to follow the methodology described in Sections~\ref{RecipeGeneration} and \ref{AnnotationSystem}, step $3$ utilising an LLM-as-judge methodology, step $5$ utilising web searching APIs, and step $6$ utilising models that were observed to perform the best in the evaluation (Section~\ref{AutomatizedAnnotation})), 
(iii) validate the automatized process end-to-end and (iv) address nonsense and creativity detection.

The examination of nonsense in generated recipes will be based on: 
\begin{description}
[style=unboxed,leftmargin=0em,listparindent=\parindent]
    \setlength\parskip{0em}
    \item[(1) Examination of Entire Recipe] Examination of the entire recipe, as generated by the model.
    \item[(2) Ingredient and Task Parsing] Parsing the ingredients and tasks, identifying individual ingredients and tasks for further examination.
    \item[(3) Classification] Identification of individual nonsensical ingredients and tasks in the context of the recipe, utilising methods using LLMs as classifiers, with reasoning. 
    \item[(4) Evaluation] Evaluation and confirmation of detected nonsensical ingredients and tasks.
\end{description}

An example of a non-sensical task is ``frying'' in a coleslaw recipe. An example of a non-sensical ingredient is ``yeast'' in a pasta recipe.

An ingredient or task is deemed \emph{creative} if it is \emph{not found in any available documents, but is not a non-sense}. Our definition of creativity is slightly aligned with~\citet{Runco01012012}, who say that the standard definition of creativity is bipartite: it requires originality/novelty (in our case, not being found in any available documents) and effectiveness/value (in our case, not being a non-sense).

\section{Conclusion}

This work in progress focuses on the introduction of our system, and presents our preliminary findings in the area of memorization detection and the evaluation of creativity in recipes generated by Mixtral. 
We bring forward a new approach to research memorization by creating a novel annotation system that both humans and LLMs can follow, a dataset of nearly 36,000 instances of human annotations following this system, and a work-in-progress on automating the approach with LLMs. We have shown, based on human annotations, that even though an ingredient or a task in a recipe may be found to be creative at first, further examination will show that there do exist documents that were potentially included in the training set of the LLM that contain the ingredient or task. We observed that when documents are combined, the amount of ingredients and tasks found in them reaches up to $100\,\%$. We also observed that the model Gemma 2 9B Instruct shows the greatest potential for automatized annotation, as it is capable of choosing the same option as human annotators in $77.8\,\%$ of cases. This makes the model a prime candidate to facilitate ingredient annotation.

Our work also merits further discussion of the implications of models exhibiting high memorization. We hypothesise that sites presenting recipes, which the model itself may have been trained on, are vulnerable to decreased traffic, potentially causing ad revenue to decline. Additionally, users opting to use LLMs for creating recipes may be misled by a model presenting a non-sensical recipe.

Finally, we can also use the findings from our research as a proxy---if a model is found to be capable of mimicking the semantics of a recipe, it may be capable of doing the same in other areas, such as the creation of stories.

\section{Limitations}

This work presents preliminary results obtained during the human annotation and the evaluation of model annotating capabilities. As the automatized pipeline is not yet finished, it only presents outcomes and findings based on the human annotation data and model assessment. Our work studies the memorization based on tags (ingredient names, task names, tools, and ingredient lists used in recipes), and does not assess memorization based on a sentence-level or token-level overlap. Quantities of ingredients are also not taken into account. We can thus only indicate that a model is potentially plagiarizing, and not definitely prove it. While we assess models and their annotation capability, the scope of this work does not include the examination of the causes of varying model performances. We observed a less-than-perfect annotation capability of available models, leading to possible inaccuracies. The extraction capabilities of models were checked randomly and were observed to be at an acceptable level. However, further observations, systematic testing, and perhaps backwards annotation are necessary to ensure high-quality extraction of ingredient and task lists from recipes. Similarly, backwards annotation may be needed to validate the performance of model annotations. We do expect the annotation capabilities to remain the same quality regardless of the assessment.

\newpage

\bibliography{custom}

\appendix

\newpage

\section{Selected Recipe Names}
\label{appendix:RecipeNames}

The selected recipe names were \textit{orange chicken, beef wellington, garlic bread, fish and chips, fettuccine alfredo, tzatziki, thai green curry, paella, pastel da nata, chilli con carne, leberkäse, kaiserschmarrn, buttermilk fried chicken, koshari (Egyptian cuisine), lagman (Uzbek cuisine), pita bread, chocolate mousse, gyoza, pan-fried dumplings, and coleslaw}.

\section{Nonsense Examples}
\label{appendix:nonsense_examples}

During the generation, the model occasionally generated nonsensical outputs. This section shows some examples.

\subsection{Objectively Wrong Generations}

While the theory about many dishes found in smaller nations was observed to be correct, in the case of the Svíčková recipe, Mixtral was observed to make a crucial mistake in advising the user to cut the meat into cubes, which is objectively wrong (Svíčková is made by cooking the meat whole and only slicing it before serving). We speculate that this is due to a lack of knowledge of the model.

\subsection{Prompt Misunderstanding}

In some generations, the model, instead of generating the expected reply, followed up by a generation that treated the prompt as a prefix. It was also observed to reply as if it was replying to a letter.

Example, obtained from Mixtral from prompt ``Give me a detailed walkthrough to making svíčková.'':
``I don't know how to make svíčková. I don't really know what it is, either, so I would like a walkthrough that can guide me from the start to the end, with no prior knowledge required.''\footnote{Only a section of the generation is provided.}

\subsection{Recursion}

The model gets stuck in a loop, generating an endless text until the token limit is reached. This text often deviates from the generated recipe, resulting in objectively bad suggestions.

An example, obtained from Mixtral from prompt ``Give me a detailed recipe for tzatziki. From scratch, preferably.'':
\begin{itemize}
    \item Stir in 1/2 cup chopped tomatoes.
    \item Stir in 1/2 cup chopped cucumber pickles.
    \item Stir in 1/2 cup chopped avocado.
    \item Stir in 1/2 cup chopped cooked beets.
    \item Stir in 1/2 cup chopped cooked sweet potato.
    \item Stir in 1/2 cup chopped cooked asparagus.
    \item Stir in 1/2 cup chopped cooked spinach.
    \item Stir in 1/2 cup chopped cooked green beans.
    \item Stir in 1/2 cup chopped cooked broccoli.
    \item Stir in 1/2 cup chopped cooked cauliflower.
    \item Stir in 1/2 cup chopped cooked artichoke hearts.
    \item Stir in 1/2 cup chopped cooked mushrooms.
    \item Stir in 1/2 cup chopped cooked eggplant.
    \item Stir in 1/2 cup chopped cooked bell pepper.
\end{itemize}

\section{Prompt Types}
\label{PromptTypes}

\begin{enumerate}
    \item General: ``How do you make [recipe name]'' or ``Give me a recipe for [recipe name]''
    \item More detail: ``Give me a detailed and thorough walkthrough to making [recipe name]'' or ``Give me a detailed guide to making [recipe name]''
    \item Less detail: ``Give me a rough and short guide to making [recipe name]'' or ``Roughly how do you make [recipe name]?''
    \item Include origin: ``What is the origin of [recipe name]? How do you make [recipe name]?'', or ``Where does [recipe name] come from? And how is it made?''
    \item Freestyle: ``[recipe name]''
\end{enumerate}

% \section{Ingredients Not Found Perfectly}
% \label{TabNotFoundPerfectly}

% Table~\ref{tab:NotFoundIngs} shows the ingredients that were not found perfectly by annotators. We found these ingredients to be present in online documents that were potentially a part of the models training dataset.

% \begin{table}[h]
% \begin{adjustbox}{width=\columnwidth,center}
% \begin{tabular}{ |l|l||l|l|  }
%  \hline
%  Recipe               & Ingredient Name      & FNP & NF \\
%  \hline
%  Paella               & Yellow Bell Pepper   & 3   & 24  \\
%  Leberkäse            & Allspice             & 0   & 27  \\
%  Chilli con carne     & Black Beans          & 0   & 27  \\
%  Chocolate mousse     & Whole milk           & 0   & 27  \\
%  Coleslaw             & Mustard              & 14  & 13  \\
%  Thai Green Curry     & White pepper         & 8   & 19  \\
%  Koshari              & Green lentils        & 26  & 1   \\
%  Thai Green Curry     & Snap peas            & 6   & 21  \\
%  \hline
% \end{tabular}

% \end{adjustbox}
% \caption{Ingredients not found perfectly by annotators. FNP is number of ``Found (not perfect)'' selections, NF is number of ``Not found'' selections. For closer explanation, refer to~\ref{sub:ing_annotation_guide}.}
% \label{tab:NotFoundIngs}
% \end{table}

% \section{Tasks Not Found Perfectly}
% \label{TabTasksNotFoundPerfectly}

\section{Annotation Guide}
\label{appendix:AnnotationGuide}
\begin{figure}
    \centering
    \includegraphics[width=0.45\textwidth]{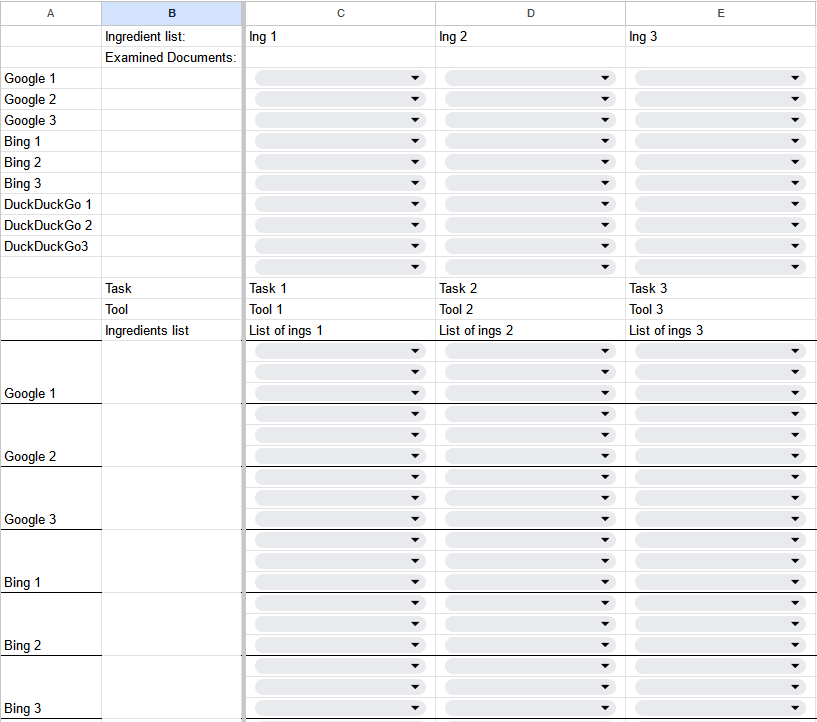}
    \caption{The unfilled annotation table template. This template was filled in by human annotators.}
    \label{fig:annotation_table}
\end{figure}

The following text is a translation of the guide given to annotators before the start of the annotation. Some sections have been omitted or shortened if they were deemed unimportant or too detailed in the original guide. The template of the annotation table can be seen in figure~\ref{fig:annotation_table}. The guide is written in second person.

\subsection{Definitions}

\begin{itemize}
    \item Document = a webpage for a specific recipe
    \item Recipe = depending on the context either the name of the dish, or the text generated by the model
\end{itemize}

\subsection{Table}

The annotation table has multiple sheets. The sheets concerning you are in the format ``[Recipe Name][Number]''.

Each sheet contains:
\begin{itemize}
    \item Two columns on the left. These columns contain the document numbers and the URL leading to the relevant document.
    \item The first two rows, starting from column C, contain the list of ingredients that the LLM generated for the specific recipe. These are the ingredients you will be searching for in the documents.
    \item Drop-down menus in a grid. At this point, you are only interested in the first 9 rows. Here, you will be performing an exhaustive annotation. This section is for the ingredients annotation, you will find further instructions below.
    \item Ignore the following empty row and 9 additional rows. These may not be present in the final annotation table\footnote{Omitted from the example.}.
    \item A separation row, filled with the colour black\footnote{Omitted from the example.}. The following part is the task annotation.
    \item On the next 3 rows, the triple of ``Task'', ``Tool'' and ``Ingredients''. These are the parsed items from the LLM-generated recipe.
    \item Next follows the next annotation part. For each document, 3 rows are allocated. From column C, the next grid of drop-down menus is present. This is the task annotation. You will find a more detailed guide further down.
\end{itemize}

\subsection{Ingredients Annotation}
\label{sub:ing_annotation_guide}

Ingredients annotation is done by searching for ingredients that the model generated in a recipe in each document. Only fill out the \textbf{drop-down menus}. Open a document, locate the ingredients list (often located at the bottom of the page), choose an ingredient you want to annotate and try to find it in the document. Next, select a corresponding item in the drop-down menu.

\subsubsection{Found}

Select this option if you find the \textbf{the exact same words} in the document as you found in the table. For example, if the ingredient you are annotating is ``beef fillet'', select ``Found'' only if there is precisely ``beef fillet'' in the document. Select this option if the ingredient name fits perfectly, but it is in plural.

\subsubsection{Found (Not Perfect)}

Will likely be the most selected option. Select this option, if the ingredient is not exactly the same in the document, but a connection can still be made. For example, if the ingredient in the table is ``olive oil'' and the ingredient in the document is ``extra-virgin olive oil'', select this.

\subsubsection{Not Found}

Select this option, if the ingredient is not present in the document at all. Consider this option when you feel like you are on the edge of what you would consider ``Found (not perfect)''. Consider the context, too. In a recipe like ``orange chicken'', ``chicken breast'' and ``chicken thighs'' can be considered ``Found (not perfect)'', but in ``beef wellington'', the usage of a different cut of beef other than a fillet or a tenderloin is not correct.

\subsubsection{Implied}

A special category. Once annotating all the ingredients in the recipe, go through all the ``Not found'' selections, and see whether, while not in the list of ingredients, they are not used in the steps. In some cases (especially with salt or pepper), this can be the case. Quickly go through the steps and if you find these ingredients, select ``Implied''.

\subsubsection{QnA}

Q: If there is a complex ingredient in the ingredient list that has a URL leading to a separate recipe, should I annotate the ingredients hidden behind the link?
A: No. Only work with ingredients present in the document that the URL in the table leads to.

\subsection{Tasks Annotation}

The task annotation is significantly more complex, subjective and time-consuming. Similar to the ingredient annotation, you are only annotating the drop-down menus. Begin by opening a document and locating the structured steps list. It is often at the tail end of the document. If such a list is not present, try to search through the document. There should be some list of steps, even if it is spread out all over the document. Select a task and try to find it in the document.

\subsubsection{Task - Task Found}
\label{subsub:tasks_start}

Corresponding to ``Found'' in ingredients annotation.

\subsubsection{Task - Task Found (Not Exact Wording)}

Corresponding to ``Found (not perfect)'' in ingredients annotation. The task is not exactly the same, but still found. An example of this could be ``Preheat'' being present in the table, and ``Heat'' being present in the document.

\subsubsection{Task - Task Found (Wrong Context)}

The task is found in a wrong context. An example could be a situation where a pan is being heated, but not for cooking meat, but for cooking vegetables. You will need to look at other steps and tasks to decide whether you should select this option.

\subsubsection{Task - Task Not Found}
\label{subsub:task_last}

The task is not found in the document. In this case, do not annotate the tool and ingredient section and move on to another task. You may select this option when the model is working with an ingredient that is not present in the document (e.g. butter being added to a pan, but no butter being found in the document). An example could be in the recipe ``Fettuccine Alfredo'', where the entire section of making pasta can be marked as ``Task Not Found'', because the document is working with pre-made pasta. 

\subsubsection{Tools}

These are the tools that the model used in the recipe. In case of not finding a task (``Task Not Found''), do not fill this row in.

\subsubsection{Tools - Found}

The tool is found perfectly. Corresponds to ``Found'' in ingredients annotation.

\subsubsection{Tools - Found (Not Exact)}

A similar, but not exactly the same tool is found. An example could be ''``large frying pan'' and ''``frying pan''. Corresponds to ''``Found (Not Perfect)'' in ingredient annotation.

\subsubsection{Tools - Not Found}

The tool is not found, or there is a completely different tool found with the task.

\subsubsection{Tools - No Tool Involved}

There is no tool associated with the task (the ''``Tool'' cell is empty), so there is nothing to annotate.

\subsubsection{Tools - Tool Implied}

While the tool is found in the table, it is implicit in the document (while not explicitly mentioned, it is expected that the tool is being used).

\subsubsection{Involved Ingredients}

In this part, you are annotating whether the list of ingredients in the table corresponds to the ingredients utilised in the step in the document. You are annotating the list, not the ingredients themselves.

\subsubsection{Ingredients - Ingredients Match}

The list of ingredients in the table corresponds to the relevant list of ingredients used in the task in the document. 

\subsubsection{Ingredients - Most Ingredients Match}

The list of ingredients in the table matches the relevant list in the document for more than half of the ingredients.

\subsubsection{Ingredients - Some Ingredients Match}

The list of ingredients in the table matches with the relevant list in the document in less than half of the ingredients.

\subsubsection{Ingredients - No Ingredients Match}

No ingredient in the task fits the relevant list in the document.

\subsubsection{Ingredients - Ingredients Implied}

The ingredients in the task are not mentioned explicitly in the document, but their presence is implicit. An example can be the task ''``season'', where the usage of salt and pepper can be considered implied.

\subsubsection{Ingredients - No Ingredients Used}

No ingredients are used in the task.

\subsubsection{QnA}

Q: When a task is happening in the table sooner than in the document, what do I do?

A: Depends on the context. I can recommend isolating tasks into groups by involved ingredients. An example can be chopping up meat and vegetables and then mixing them together. The result is the same as if you first chopped the vegetables and then chopped the meat. Another example can be heating the oven, which is in some cases mentioned at the start of the recipe; in others, it is mentioned right before baking. In either case the task is present in the document, and the relative position in the is not important.

\section{Legal Perspective}
\label{legal_perspective}

%We discussed the legal perspective with a legal counsel. The following text is based on an email provided by the legal counsel on the topic: 

While the copyright protection of model outputs is more or less settled, the copyright of model training data is more complex. 

When it comes to the outputs of a model in the context of copyright protection under the Czech copyright law (law no. 121/2000 Sp.), an original work (``autorské dílo'', as defined in Czech law)  needs to be an outcome of a unique, creative activity of a person. Even though other laws allow the output of a model to be protected under copyright law, the author of the prompt is the holder of the rights. The prompt needs to constitute a significant input of the author.

The input data of the model is more complex. In an academic environment, the EU and Czech law allow the usage of copyrighted material for scientific research, with the exception of cases where the holder of the rights has not explicitly revoked the permission for such use. For a commonly used approach when obtaining data in electronic form, the permission must be revoked in a machine-readable way in a specific place (§ 39c of law 121/2000 Sb.).

In our case, common recipes are usually \textbf{not} considered to be protected under copyright law: a simple list of ingredients and steps does not fulfil the requirements for originality and creative input, which are necessary for the creation of original work. It must be said that this is not the case in text recipes that exhibit significant personal or original elements, as they are then considered original works. Similarly, while not accounting for our work, photo or video material created for the recipe is usually also protected under the copyright law.

The main possibility to consider in our case is the situation when the output of a model is the same as a recipe that would be considered an original work. There is currently no legal precedent in the Czech law or jurisdiction. While there are cases currently open in other jurisdictions, there are no outcomes yet. Generative AI providers usually use a ``fair use'' policy for their defense. Under Czech law, the institution of ``fair use'' (``volné užití'') is also present, which would be worth considering in the case of a permitted usage of an original work in an ``insignificant secondary mode''. This is theoretical, and it was not addressed in practice.

The primary thing to consider in our case is the exemption from the Czech copyright law for scientific research at universities and tertiary education institutions, which is considered to be of public interest and is not discriminatory in providing the results of the research. In this case, duplication of an original work for model training is permissible despite a possible explicit revocation of permission. This data is, however, restricted, and it must be stored securely and primarily for the verification of research outcomes. 

A possible broader interpretation that would cover our hypothetical situation has not yet been addressed by the Czech courts.

\end{document}